\pdfoutput=1
\documentclass[11pt]{article}

\usepackage{acl}

\usepackage{times}
\usepackage{latexsym}

\usepackage[T1]{fontenc}
\usepackage[utf8]{inputenc}

\usepackage{microtype}

\usepackage{booktabs} % For formal tables
\usepackage{graphicx} 
\usepackage[caption=false]{subfig}
\usepackage{color}
\usepackage{algorithm}
\usepackage{algorithmic}

\usepackage{microtype} 
\usepackage{amsmath,amsfonts}
\usepackage{mathtools} 
\usepackage{mathrsfs}  
\usepackage{bm}
\usepackage{booktabs, multirow}
\usepackage{makecell}
\usepackage{stfloats}
\newcommand{\bluehl}[1]{{\color{blue}#1}}

\usepackage{CJKutf8}
\usepackage{ulem} 
\usepackage{arydshln}
\usepackage{enumerate}

\usepackage[subtle]{savetrees}

\title{\textsc{Creater}: CTR-driven Advertising Text Generation with\\ Controlled Pre-Training and Contrastive Fine-Tuning}

\author{Penghui Wei, Xuanhua Yang, Shaoguo Liu\thanks{\hspace{1mm} Corresponding author.}, Liang Wang \and Bo Zheng \\
         Alibaba Group \\ {\{wph242967,xuanhua.yxh,shaoguo.lsg,liangbo.wl,bozheng\}@alibaba-inc.com}}

\begin{document}
\maketitle
\begin{abstract}
This paper focuses on automatically generating the text of an ad, and the goal is that the generated text can capture user interest for achieving higher click-through rate (CTR). We propose \textsc{Creater},\footnote{\textbf{C}T\textbf{R}-driv\textbf{E}n \textbf{A}dvertising \textbf{TE}xt Gene\textbf{R}ation} a CTR-driven advertising text generation approach, to generate ad texts based on high-quality user reviews. To incorporate CTR objective, our model learns from online A/B test data with contrastive learning, which encourages the model to generate ad texts that obtain higher CTR. To alleviate the low-resource issue, we design a customized self-supervised objective reducing the gap between pre-training and fine-tuning. Experiments on industrial datasets show that \textsc{Creater} significantly outperforms current approaches. It has been deployed online in a leading advertising platform and brings uplift on core online metrics.
\end{abstract}

\section{Introduction}\label{intro}
\begin{figure}[t]
\centering
\centerline{\includegraphics[width=0.95\columnwidth]{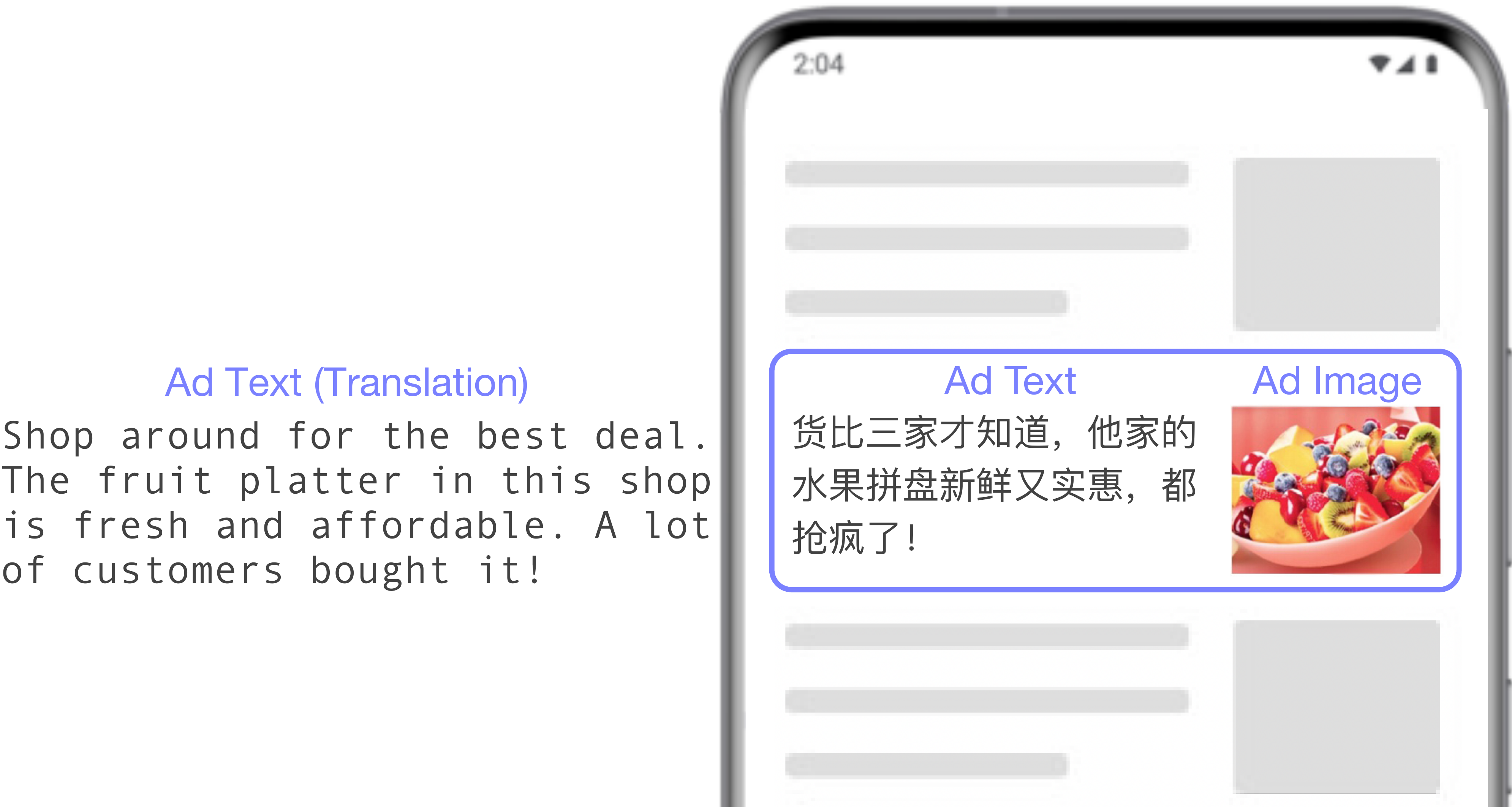}}
\caption{An illustration that shows the creative of an online advertisement in news feed on mobile.}
\label{fig:example}
\end{figure}

For businesses that want to promote their items and services, running online advertisements on advertising platforms is an effective way to achieve their marketing goals. 
With the aim of attracting users to know more about the displayed items, advertisers design ad creative (such as text, image and video). Figure~\ref{fig:example} is an illustration that shows the creative of an ad in news feed, which contains a text and an image.

An appropriate creative design capturing user interest accurately can improve the ad's click-through rate (CTR). CTR is a key metric that quantifies the effect of an ad, because click is the precondition for any further actions such as sharing and purchase taken by users. Thus designing ad creatives that can achieve higher CTR is crucial for ad delivery. 

Traditionally, advertisers need to manually design the creative of each ad, and then resort to online A/B test results to continually refine initial creative for catching user interests. Such trail-and-error process is labor-intensive and usually inefficient. 
In terms of the text in a creative, due to the variation characteristic of language expressions, it may need to be polished multiple times for obtaining an ideal one. 
To improve the efficiency of ad delivery for advertisers, especially for small advertisers that may not afford to hire professional writers, this paper focuses on automatically generating the text for an ad, and the goal is that the generated text can capture user interest for achieving higher CTR. 

There are several challenges to achieve this goal. 
\textbf{(I)} First, it is important to choose a suitable source for generating ad texts. A straightforward source is the corresponding item's title in landing page. However, a title is usually a mixture of item attributes while may not reflect user preference. In contrast, an ad text should contain insightful and informative contents that can arouse purchasing desire of users. 
\textbf{(II)} Second, most of current natural language generation (NLG) models are optimized using cross-entropy criterion, which is discrepant to the CTR metric we concern. To encourage the model to generate texts achieving higher CTR, there is a great need to incorporate CTR objective into training. 
\textbf{(III)} Last but not least, a well-trained NLG model usually need a large amount of paired data. However it is costly to collect sufficient human-written ad texts, especially for small advertisers, thus we are faced to low-resource problem. 

In this paper we propose \textsc{Creater}, a CTR-driven advertising text generation approach, to address the above challenges. 
\textbf{(I)} First, we choose high-quality user reviews as input source for generation. Compared to titles, user reviews intuitively contain contents that reflect real experience after purchasing. We also introduce an aspect term as input control code to improve the informativeness of generated text.
\textbf{(II)} Second, to explicitly incorporate CTR objective during optimizing NLG models, we make use of collected user feedback through online A/B test. Advertisers always perform online A/B test to compare two different texts of a same ad, where online CTR metric reflects the distinction between a relatively ``good'' text and a ``bad'' one. We employ contrastive learning for model optimization, which encourages our model to generate texts that can achieve high CTR. 
\textbf{(III)} Finally, to alleviate the low-resource problem, we make use of large-scale unpaired reviews to perform pre-training that provides warm-starting. We design a novel self-supervised objective customized to our scenario, which reduces the gap between pre-training and fine-tuning.

\textsc{Creater} has been deployed online in a leading advertising platform and it achieves significant improvement on core online metrics. 
The main contributions of this work are summarized as follows:

$\bullet$ We propose \textsc{Creater} for generating ad texts that capture user interest based on high-quality user reviews. We make use of online A/B test data to perform contrastive learning, which encourages the model to generate texts that achieve higher CTR.

$\bullet$ We propose a novel self-supervised objective to provide warm-starting with unpaired reviews, which is customized to our scenario and reduces the gap between pre-training and fine-tuning. 

$\bullet$ Experiments on industrial datasets show that \textsc{Creater} outperforms previous approaches on both automatic and human evaluation, and online results verify that it brings significant uplift on core metrics.
%\end{itemize}

\begin{figure*}[t]
\centering
\centerline{\includegraphics[width=2\columnwidth]{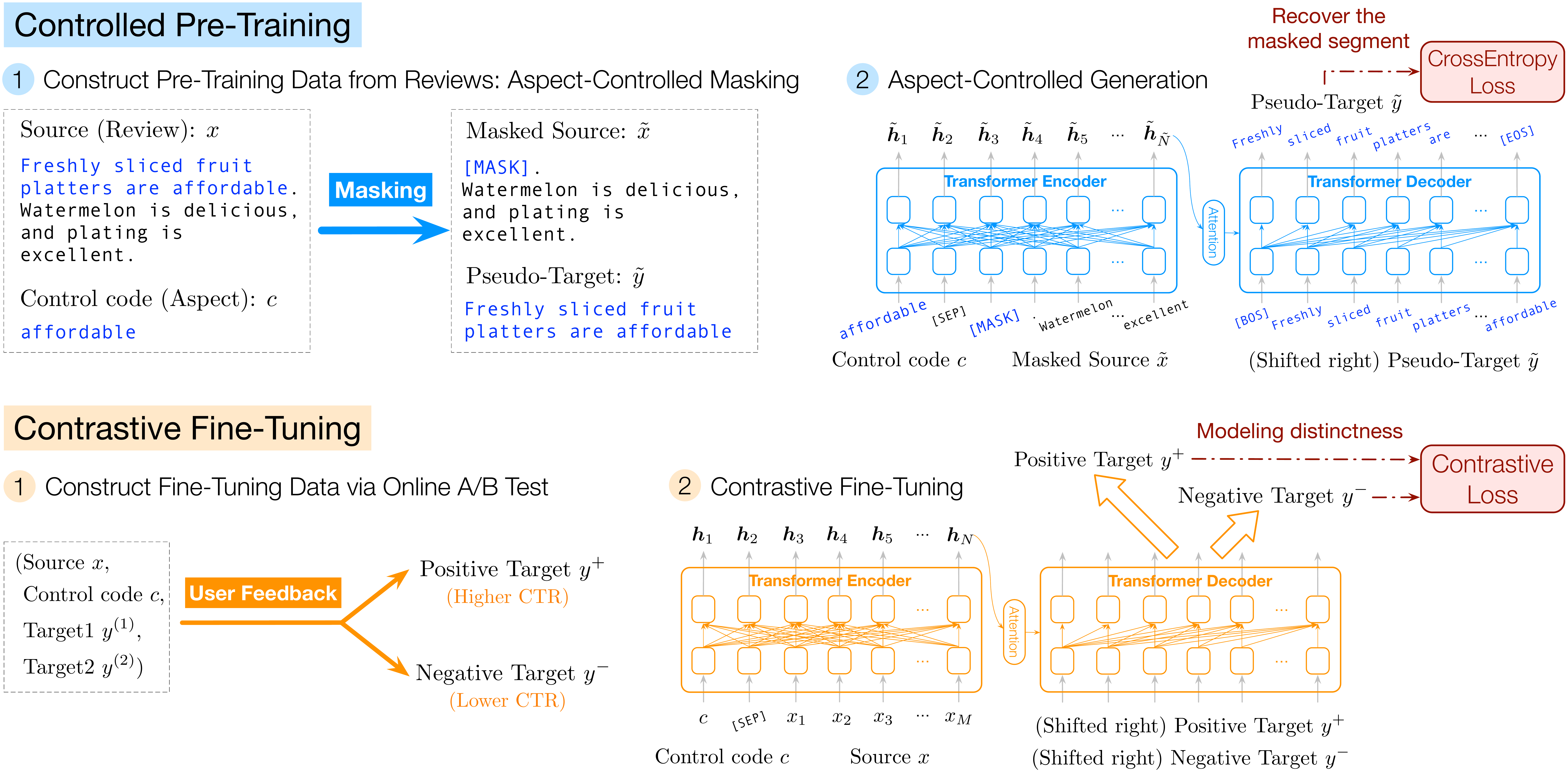}}
\caption{Overview of our proposed approach \textsc{Creater} for CTR-driven advertising text generation. }
\label{fig:overview}
\end{figure*}

\section{Problem Formulation}\label{method:formulation}

Given a source $x$ and a control code $c$ for an ad, where the source is a high-quality user review of the ad item, the control code is an aspect term of such review to guide generation, we aim to learn a generation model $p_{\Theta}(y\mid x,\ c)$ that can produce an appropriate ad text $y$ (where $\Theta$ denotes trainable parameters of the model). Our goal is that the generated ad text can capture user interest and attract users to know more about the ad item.

\section{Proposed Approach: \textsc{Creater}}
Figure~\ref{fig:overview} illustrates the workflow of our \textsc{Creater}, and it consists of two stages. The first stage is \textit{Controlled Pre-Training}, which learns from unpaired user reviews to provide warm-starting for low-resource scenario. The second stage is \textit{Contrastive Fine-Tuning}, which further learns from online A/B test data that reflects user feedback, aiming to encourage the model to generate ad text that can achieve higher CTR.

\subsection{Stage 1: Controlled Pre-Training}\label{method:stage1}
We construct a large set of user reviews as the pre-training corpus $\mathcal D_x$. 
Based on $\mathcal D_x$, we extract a set of aspect terms $\mathcal D_c$ using an off-the-shelf unsupervised model \textsc{Abae}~\cite{he2017unsupervised}, and each aspect term is typically represented as a word. 

Recall that we aim to learn a generation model $p_{\Theta}(y\mid x,\ c)$, while the pre-training stage only makes use of \textit{unpaired} user reviews $\mathcal D_x$. To ensure that the model benefits from pre-training, we propose a novel self-supervised objective customized to our scenario, which reduces the gap between pre-training and fine-tuning.
The core is that, for each review $x\in\mathcal D_x$, we construct an aspect-based \textit{pseudo-target} $\tilde y$ from the review $x$ and mask this segment in $x$. The self-supervised objective is to perform aspect-controlled generation, which aims to recover the segment $\tilde y$ given the masked review with the guidance of corresponding aspect term. 

\paragraph{Aspect-Controlled Masking}\label{method:stage1_masking}
For a review $x\in\mathcal D_x$, we tokenize it as a list of segments $[x_\mathrm{seg\_1}, x_\mathrm{seg\_2}, \ldots]$ based on punctuations and dependency parser, where each segment $ x_\mathrm{seg\_i}$ is a sub-sequence of $x$. 
Given an aspect term $c\in\mathcal D_c$ existed in the review $x$, we compute the matching score  between $c$ and each $ x_\mathrm{seg\_i}$ with a matching function $f\left(c,\ x_\mathrm{seg\_i}\right)$.\footnote{The function $f(\cdot, \cdot)$ can either be a lexical-based one (such as similarity of sparse TF-IDF vectors) or an embedding-based one (such as similarity of averaged word embeddings).} 
We then select the segment with highest matching score as the \textit{pseudo-target} ${\tilde y}$ for the given pair of (source $x$, control code $c$):
\begin{equation}
\label{method:matching}
     {\tilde y}=\arg\max\limits_{x_\mathrm{seg\_i}\ \in\ x} f\left(c,\ x_\mathrm{seg\_i}\right)
\end{equation}

For each triple (source $x$, control code $c$, pseudo-target $\tilde y$), our aspect-controlled masking strategy masks the review $x$ by replacing its pseudo-target $\tilde y$  with a special word ``\texttt{[MASK]}''. Thus we transform each triple $(x,\ c,\ \tilde y)$ to a masked one $(\tilde x,\ c,\ \tilde y)$, where the masked review $\tilde x$ is specific to the aspect term $c$.

\paragraph{\textbf{Aspect-Controlled Generation}}
Given a masked review $\tilde x$ with an aspect term $c$, our self-supervised objective is to recover the masked segment (i.e., pseudo-target $\tilde y$) of original review $x$ with the controlling of $c$:
\begin{equation}\label{method:stage1-objective}
    \min_{\Theta}\ -\log\ p_{\Theta}\left(\tilde y\mid \tilde x,\ c\right)\,.
\end{equation}

Such aspect-controlled generation enforces the model to understand the context of input masked review better. Compared to general pretraining models~\cite{zhang20pegasus,lewis2020bart,raffel2020exploring}, the proposed objective is customized to our scenario. The input information $\tilde x$ does not contain the content to be generated, improving the ability of generating abstractive contents other than simply copying from input only. 

Formally, we first prepend the control code $c$ to the masked source $\tilde x$, and add a special word \texttt{``[SEP]''} between them. We then feed the concatenated sequence $[c,  {\small\texttt{[SEP]}},\ \tilde x]$ into \textsc{Creater} to generate the pseudo-target $\tilde y=[\tilde y_1,\tilde y_2,\ldots,\tilde y_{\tilde T}]$ (where $\tilde T$ denotes the length), where the model architecture is a Transformer encoder-decoder~\cite{vaswani2017attention} and it is optimized via teacher-forcing: 
\begin{equation}
\begin{aligned}
    \bm{{\tilde h}}_1, \bm{{\tilde h}}_2, \ldots,  \bm{{\tilde h}}_{\tilde N}\ =\ \mathsf{Enc}\left([c,  {\small\texttt{[SEP]}},\ \tilde x]\right) \\
    p\left(\tilde y_t\mid \tilde y_{0:t-1},\ \tilde x,\ c\right) \sim \mathsf{Dec}\left(\tilde y_{0:t-1},\ \bm{{\tilde h}}_{1:{\tilde N}} \right)\\
    \min_{\Theta} \sum_{t=1}^{\tilde T} -\log\ p_{\Theta}\left(\tilde y_t\mid \tilde y_{0:t-1},\ \tilde x,\ c\right)
\end{aligned}
\end{equation}
where $\tilde N$ is the length of the sequence $[c,  {\small\texttt{[SEP]}},\ \tilde x]$, and $\bm{{\tilde h}}_i$ is the $i$-th word's representation.

\subsection{Stage 2: Contrastive Fine-Tuning}\label{method:stage2}
To incorporate CTR objective during generation, we make use of existing online A/B test data that reflects user preference. Specifically, we construct a dataset $\mathcal D$, where each sample is a tuple (source $x$, control code $c$, positive target $y^+$, negative target $y^-$). Both $y^+$ and $y^-$ are human-written ad texts (given $x$ and $c$), while $y^+$ achieves higher CTR than $y^-$ during online A/B test.

Next, we start from describing a vanilla fine-tuning objective that only considers $y^+$. We then introduce two contrastive fine-tuning objectives which take good advantage of online A/B test data. 

\paragraph{Vanilla Fine-Tuning} 
A straightforward objective is to maximize the generation probability of positive target $y^+$:
\begin{equation}
    \mathcal L_{ft}\ = \ -\log\ p_{\Theta}(y^+ \mid x,\ c)\,.
\end{equation} 
Obviously, this learning objective omits the utility of negative targets. 

To enhance the model's discriminative ability of ad texts with different CTR, we propose to expose the decoder to both positive and negative ad texts via modeling their distinctness. Specifically, we leverage the paradigm of contrastive learning, where the positive/negative target (i.e., ad text with higher/lower CTR) is used to construct positive/negative paired instance, and introduce two contrastive learning based objectives to fine-tune the pre-trained model. 

\paragraph{i. Margin-based Contrastive Fine-Tuning}
We first propose to directly maximize the margin of generation probabilities between the positive target $y^+$ and the negative target $y^-$. This yields the following loss function:
\begin{equation}\label{method:margin}
\small
   \mathcal L_{cont}= \max\left\{0, -\left(\log\ p_{\Theta}(y^+ \mid x,\ c) - \log\ p_{\Theta}(y^- \mid x,\ c)\right) +\gamma \right\}
\end{equation}
where the margin $\gamma$ is a hyperparameter. Through this loss, the optimization procedure is encouraged to maximize the probability gap of ad texts having distinct CTR. 

\paragraph{ii. InfoNCE-based Contrastive Fine-Tuning}
From the perspective of representation learning, we propose a contrastive loss based on InfoNCE~\cite{oord2018representation}, which maximizes the similarity between source and positive target, and minimizes that between source and negative target: 
\begin{equation}\label{method:infonce}
\small
\mathcal L_{cont}=- \log\frac{ \exp\left({\mathrm{sim}\left((c,x),\ y^+\right)}/{\tau}\right) }{ \exp\left({\mathrm{sim}\left((c,x),\ y^+\right)}/{\tau}\right) + \exp\left({\mathrm{sim}\left((c,x),\ y^-\right)}/{\tau}\right) }
\end{equation}
where $\tau$ is temperature. $\mathrm{sim}(\cdot, \cdot)$ is similarity function of encoder and decoder representations. 

We adopt mean-pooling to the top layer of the encoder/decoder as their representations. Let $\bm h$, ${\bm z}^+$ and ${\bm z}^-$ denote encoder representation, decoder representations for positive and negative targets. 
We then add two fully-connected layers to the encoder and the decoder side respectively, transforming them to the same vector space. Thus an inner product operation is used to obtain the similarity scores:
\begin{equation}
\small
    \begin{aligned}
\operatorname{sim}\left((c, x),\ y^{+}\right) &=\left(\mathbf{W}_{e} \boldsymbol{h}\right)^{\top}\left(\mathbf{W}_{d} \boldsymbol{z}^{+}\right) \\
\operatorname{sim}\left((c, x),\ y^{-}\right) &=\left(\mathbf{W}_{e} \boldsymbol{h}\right)^{\top}\left(\mathbf{W}_{d} \boldsymbol{z}^{-}\right)\
\end{aligned}
\end{equation}
where $\mathbf{W}_{e}$ and $\mathbf{W}_{d}$ learnable parameters.

\paragraph{\textbf{Objective}} 
The final loss of contrastive fine-tuning stage is the sum of $ \mathcal{L}_{ft}$ and contrastive loss:
\begin{equation}
    \mathcal{L}_{ft}(y^+)\ + \mathcal{L}_{ft}(y^-)\ +\ \alpha \mathcal{L}_{cont}
\end{equation}
where $\alpha$ is a trade-off hyperparameter, and $\mathcal{L}_{cont}$ can either be margin-based or InfoNCE-based. 

\paragraph{\textbf{Comparison}} 
The advantage of margin-based loss is that it does not add extra parameters, directly incorporating CTR objective to generation probabilities. InfoNCE-based loss considers encoder representations to learn better decoder representations. Although it adds a few parameters (i.e., two fully-connected layers), they are pruned at inference. 
The construction of positive-negative pairs in \textsc{Creater} is designed for CTR objective via user feedback, unlike recent work tackling other issues~\cite{cao2021cliff,pan2021contrastive,lee2021contrastive}

\section{Experiments}

\subsection{Experimental Setup}

\paragraph{\textbf{Datasets}}\label{experiment:dataset}
To our knowledge, there is no available public dataset that contains ad texts coupled with CTR information, thus we collected data on a leading advertising platform.  
We construct a dataset $\mathcal D$ where each sample is a tuple of (user review, aspect term, positive ad text$_1$, negative ad text$_2$), in Chinese, through online A/B test.  
Overall the user reviews are ensured to be high-quality based on rules and filtering models. 
Each ad text is written by human editors given the review and aspect term, covering 4,047 advertisers. 
More details about data preprocessing and filtering can be found in \textbf{Appendix}~\ref{appendix:dataset}. 

We also produce a large-scale review corpus $\mathcal D_x$ for constructing pre-training dataset via aspect-controlled masking (§~\ref{method:stage1_masking}). 
Table~\ref{table:datasets} lists the statistics. We split $\mathcal D$ with 7:1:2 to obtain the training/development/test set.

\begin{table}[t]
\centering
\scriptsize
\begin{tabular}{ccc}
\toprule
\textbf{Dataset} & Pre-training ($\mathcal D_x$) & Fine-tuning ($\mathcal D$)\\
\midrule
\# Samples  & 1,471,106 & 43,985 \\ 
Avg. length of reviews & 25.05 & 25.31\\
Avg. length of ad texts & N/A & 13.06\\
\bottomrule
\end{tabular}
\caption{Statistics of the datasets used in our experiments. ``Avg. length'' means the average number of characters in a sequence (review or ad text).} 
\label{table:datasets}
\end{table}

\paragraph{\textbf{Comparative Approaches}}
We choose two types of comparative approaches in our experiments. 
The first type contains \textit{non-CTR-driven approaches}: 
(1) {\underline{\textsc{SegExt}}}  (Segment extraction)\quad employs unsupervised aspect-controlled masking (§~\ref{method:stage1_masking}) to return a segment of source as the ad text. {If the returned segment is too short to display, we add its left or right segment based on matching score.} 
(2) {\underline{\textsc{PGNet}}} (Pointer-generator)\quad  is an  RNN-based approach via copying mechanism~\cite{see2017get}; 
(3) {\underline{\textsc{C-PGNet}}}\quad improves \textsc{PGNet} by adding control code during decoding, which imposes on the generation gate; 
(4) {\underline{\textsc{Trm}}} (Transformer)\quad is the state-of-the-art architecture for text generation; 
(5) {\underline{\textsc{C-Trm}}}\quad improves \textsc{Trm} by adding control code at both encoder and decoder sides, with the help of fusion layers; 
(6) {\underline{\textsc{C-Trm-RL}}}\quad fine-tunes the \textsc{C-Trm} with reinforcement learning (RL), where an extra CTR regression model (trained on $\mathcal D$) is the reward estimator that produces click probability of a generated text~\cite{hughes2019generating}. {Negative targets are used to train the reward estimator, and are not explicitly used for optimizing generation model.}

The second type contains \textit{CTR-driven approaches}. They exploit negative target $y^-$ during training to explicitly incorporate CTR information: (1) {\underline{\textsc{QualityModel}}}\quad employs click behavior as a quality measure for paired samples~\cite{wang2019quality}. It first builds a CTR latent space to represent source and target, and then computes the cosine similarity between them as the quality score of the sample. Quality scores are used to weight the cross-entropy objective and reduce the probability of generating low-quality texts; 
(2) {\underline{\textsc{ContraModel}}} \quad is a variant of \textsc{Creater}, which removes the controlled pre-training stage; 
(3) {\underline{\textsc{Bart+ContraModel}}}\quad performs pre-training from scratch using the self-supervised objective of \textsc{Bart} other than our proposed one, and then performs fine-tuning with \textsc{ContraModel}. 

\subsection{Implementation Details}\label{appendix:implementation}
Both the encoder and the decoder of \textsc{Creater} contain four layers, and the dimension of hidden representations produced by each layer is set to 512.  
For fair comparison, all comparative approaches that based on Transformer employ the above architecture. 
For text preprocessing, we tokenize sources and targets to word sequences, and thus our \textsc{Creater} generates ad texts at word-level. We restrict the max length of input as 128 words. The overall parameter size is 129M. At the pre-training stage, we employ Adafactor optimizer~\cite{shazeer2018adafactor}, with a mini-batch size of 4096 for training 10 epochs. Models are trained on 8 Tesla V100 32GB GPUs. We implement our approach with \textit{PyTorch}\footnote{\url{https://github.com/pytorch/pytorch}} and \textit{Transformers}\footnote{\url{https://github.com/huggingface/transformers}}.

In terms of the model for extracting aspect term set, during early experiments we found that the performance of \textsc{Creater} is not sensitive to it and thus we employ the representative model \textsc{Abae}. 
For matching function $f(\cdot, \cdot)$ (Equation~\ref{method:matching}) used in aspect-controlled masking for building pre-training data, we try a lexical-based (similarity of sparse TF-IDF vectors) and an embedding-based one (similarity of averaged word embeddings), and found that the performance of fine-tuned model is not sensitive to them. Thus we choose the former for simplicity. 

At the fine-tuning stage, we set the mini-batch size to 1024 for 20 epochs. When the margin-based contrastive loss is used, the margin parameter $\gamma$ is set to 1.0. Or if we the use InfoNCE-based contrastive loss, the temperature parameter $\tau$ is set to 1.0. We set the trade-off hyperparameter $\alpha$ to 1e-3 (which is searched from \{1e-2, 1e-3, 1e-5\}). We choose the checkpoint that has lowest perplexity on validation set as the final model.  
At inference time, we use beam search algorithm to generate texts, where the beam size is set to 5. 
The BLEU metric is evaluated using \textit{NLTK}\footnote{\url{ https://github.com/nltk/nltk}}, and the ROUGE metric is evaluated using \textit{pyrouge}\footnote{\url{https://github.com/bheinzerling/pyrouge}}. All reported results of different approaches are run based on the same random seed.

\begin{table}[t]
\centering
\scriptsize
\begin{tabular}{lcccc}
\toprule
\textbf{Approach} & BLEU-4 & RG-1 & RG-2 & RG-L \\ 
\midrule
\multicolumn{5}{l}{\textit{Non-CTR-driven Approaches}}\\
\textsc{SegExt} & 13.54 & 31.11 & 7.71 & 23.66 \\
\textsc{PGNet}  & 24.85 & 44.79 & 16.76 & 35.21\\
\textsc{C-PGNet} & 37.69 & 55.09 & 31.70 & 46.62\\
\textsc{Trm}  & 33.36 & 50.58 & 26.23 & 42.44 \\
\textsc{C-Trm} & 48.66 & 61.73 & 42.43  & 54.82\\
\textsc{C-Trm-RL} & 50.11 & 62.59 & 42.26 & 55.43\\
\midrule
\multicolumn{5}{l}{\textit{CTR-driven Approaches}}\\
\textsc{QualityModel} & 49.89 & 62.67 & 43.85 & 55.84\\
\hdashline[2pt/1.2pt]
\textsc{ContraModel} & {51.47} & {63.47} & {43.94} & {56.93} \\
\textsc{Bart+ContraModel} & 53.35 & 65.04 & 46.20 & 58.51\\
\textsc{Creater}  & \textbf{54.56} & \textbf{65.93} & \textbf{47.44} & \textbf{59.77}\\
\bottomrule
\end{tabular}
\caption{Main results. ``RG'' stands for ROUGE.   Both BLEU and ROUGE scores are multiplied by 100.}
\label{results:main}
\end{table}

\subsection{Performance Comparison}
Table~\ref{results:main} shows the comparison results, and we report BLEU-4 and ROUGE-1/2/L (positive targets are regarded as gold-standard).\footnote{Our \textsc{Creater} performs significantly better than the second best comparative approach at the level of $p<0.05$.} 
It is natural that the approaches considering aspect terms outperform those that do not perform controlling.

CTR-driven approaches usually outperforms non-CTR-driven ones, demonstrating that exposing the model to both positive and negative targets improves generation quality. 
\textsc{QualityModel} and \textsc{ContraModel} represent two paradigms to incorporate CTR information. \textsc{ContraModel} is superior to \textsc{QualityModel}, which indicates that directly modeling the distinctness as an auxiliary objective is more effective than weighting the original loss. 

\textsc{Bart+ContraModel} performs better than \textsc{ContraModel} by adding a pre-training stage. 
\textsc{Creater} proposes a customized controlled pre-training objective and achieves the best result. This verifies that designing a suitable self-supervised objective is crucial to improve generation. 

\begin{table}[t]
\centering
\scriptsize
\begin{tabular}{lcccc}
\toprule
\textbf{Variants of Pre-training} & BLEU-4 & RG-1 & RG-2 & RG-L \\ 
\midrule
{\textsc{Creater}} ($p(\tilde y\mid \tilde x,c)$) & \textbf{54.56} & \textbf{65.93} & \textbf{47.44} & \textbf{59.77} \\
{\quad w/o masking}  ($p(\tilde y\mid  x,c)$) & 51.24 & 63.65 & 43.74 & 56.94\\
{\quad w/o control code}  ($p(\tilde y\mid \tilde x)$) & 53.11 & 64.64 & 45.91 & 58.28\\
{\quad w/o whole pre-training} & 49.92 & 62.20 & 41.91 & 55.09 \\
\bottomrule
\end{tabular}
\caption{Comparison of  pre-training objectives. }
\label{results:discussion_stage1}
\end{table}

\subsection{Discussion}

\paragraph{Effect of Aspect-Controlled Masking}
During pre-training, aspect-controlled masking ensures the ability of generating abstractive contents other than simply copying from source. Besides, the model takes aspect terms as control codes to generated masked contents (pseudo-targets). Both the two mechanisms reduce the gap between pre-training and fine-tuning. 
We verify their effectiveness by removing one of two mechanisms, and the fine-tuning stage keeps unchanged. 
Results are shown in Table~\ref{results:discussion_stage1}. The two variants are inferior to the full model, demonstrating that both of them can improve pre-training to provide better warm-starting. Aspect-controlled masking brings improvements over 3 BLEU score and 2 ROUGE score. Thus, our novel controlled pre-training objective indeed enhances the performance of advertising text generation via effective self-supervised learning on unpaired corpus.

\begin{figure}[t]
\centering
\centerline{\includegraphics[width=0.65\columnwidth]{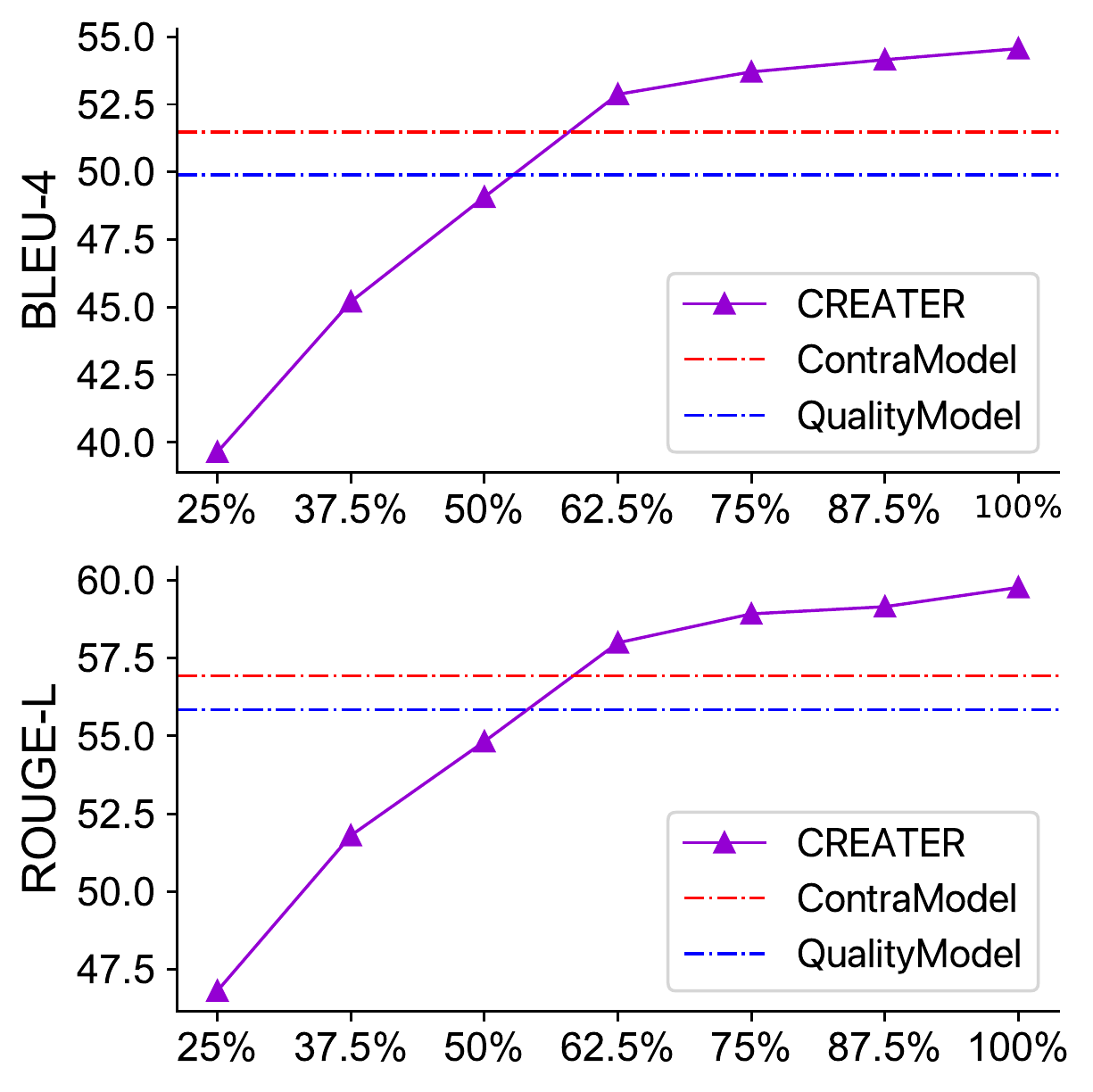}}
\caption{Results with limited fine-tuning data. Dashed lines are two strongest baselines trained on whole data.}
\label{fig:lowresource}
\end{figure}

\paragraph{Benefit in Low-Resource Scenario}
We further verify the effect of controlled pre-training when there are only limited paired data for fine-tuning. We change the size of data (from 25\% to 100\% of the whole training set), and compare to two strongest baselines (\textsc{QualityModel} and \textsc{ContraModel}, without pre-training) that are trained on the whole training set. As shown in Figure~\ref{fig:lowresource}, with only half of fine-tuning data, \textsc{Creater} performs on par with \textsc{QualityModel}, verifying the benefit of our controlled pre-training in low-resource scenario.

\paragraph{Analysis of Contrastive Fine-Tuning}
Our \textsc{Creater} exposes the model to both positive and negative targets for incorporating CTR information. 
Table~\ref{results:discussion_stage2} shows the comparison of two contrastive objectives. For with and without pre-training, the best-performing model is based on contrastive learning. 

An interesting point is that when we perform pre-training, InfoNCE-based model achieves best performance, while margin-based model outperforms other variants if we do not pre-train the model. We suggest that InfoNCE-based loss is designed from the perspective of representation learning, and pre-training can provide better text representations compared to no pre-training. Thus in this situation the utility of InfoNCE-based model is highlighted.

\begin{table}[t]
\centering
\scriptsize
\begin{tabular}{cccccc}
\toprule
\multicolumn{2}{c}{\textbf{Variants of Contrastive Loss}} & \multirow{2}*{BLEU-4} & \multirow{2}*{RG-1} & \multirow{2}*{RG-2} & \multirow{2}*{RG-L} \\
\cmidrule(lr){1-2}
Pre-Train & Contrastive Loss & \\
\midrule
\checkmark & InfoNCE-based & \textbf{54.56} & \textbf{65.93} & \textbf{47.44} & \textbf{59.77} \\
\checkmark & Margin-based &  54.26 & 65.93 & 47.23 & 59.57 \\
\checkmark & No & 53.70 & 65.38 & 46.57 & 58.94 \\
\cmidrule(lr){1-6}
$\times$ & InfoNCE-based &  49.92 & 62.20 & 41.91 & 55.09 \\
$\times$ & Margin-based  & \textbf{51.47} & \textbf{63.47} & \textbf{43.94} & \textbf{56.93}  \\
$\times$ &  No & 50.37 & 62.27 & 42.19 & 55.36 \\
\bottomrule
\end{tabular}
\caption{Comparison of contrastive learning objectives. }
\label{results:discussion_stage2}
\end{table}

\begin{table}[t]
\centering
\scriptsize
\begin{tabular}{ccccc}
\toprule
\textbf{Approach} & Gram. & Info. & Suit. & Avg. Rank ($\downarrow$) \\  
\midrule
\textsc{SegExt}  & \textbf{4.97} & 2.19 & 1.92 & 4.53\\
\textsc{C-Trm} & 4.95 & 2.69 & 2.44 & 3.65\\
\textsc{QualityModel} & 4.96 & 2.81 & 2.49 & 3.19 \\
\textsc{Creater}  & 4.96 & \textbf{3.21} & \textbf{3.05} & \textbf{2.09}\\
\cmidrule(lr){1-5}
Human-written (high-quality) & 4.99 & 3.60 & 3.22 &1.48\\
\bottomrule
\end{tabular}
\caption{Human evaluation results. 
``Gram.'', ``Info.'', ``Suit.'' and ``Avg. Rank'' stand for grammaticality, informativeness, suitability and average rank, respectively. }
\label{results:human}
\end{table}

\subsection{Human Evaluation}\label{experiment:human}
An ad text will be measured from the three views: grammaticality, informativeness (whether its content reflects the key points of aspect term and the source) and suitability (whether it is suitable to be displayed).
Each view is ranging from 1 to 5 (5 is the best). We randomly choose fifty samples and invite three human judgments.

Table~\ref{results:human} shows that
\textsc{Creater} performs well on most views and achieves the best ranking results among four comparative approaches,  possessing the ability of generating fluent, informative and suitable ad texts. 
We found that the reason why the informativeness and suitability of \textsc{Creater} are not as high as human-written ones is that the faithfulness of generated texts is not always ideal. We leave the improvement in future work.

\subsection{Case Analysis}
We further show the generated ad texts from different approaches for case analysis. 
Table~\ref{results:case} is a case analysis that the input contains a source review with an aspect term. By comparing these generated results, 
We can see that the ad text generated by \textsc{Creater} is more suitable to attract users. 
The generated phrase \texttt{``sweet, quenching your thirst''} is more attractive than other results like \texttt{``tastes well''}.
On the whole, the overall quality of the ad texts generated by \textsc{Creater} is better than other competitive approaches.

\begin{table}[t]
\centering
\scriptsize
\begin{tabular}{p{6em}p{23em}}
\toprule
\multirow{2}*{\textbf{Approach}} & {\textbf{Source}: \begin{CJK*}{UTF8}{gbsn}水果很新鲜, 口感很好吃着非常甜, 价格优惠, 下次还会光顾\end{CJK*} (The fruit is fresh, and it tastes delicious and sweet. The price is favorable. Will buy it next time.)}\\
& \textbf{Control code}: \begin{CJK*}{UTF8}{gbsn}口感\end{CJK*} (taste) \\  
\midrule 
\textsc{SegExt}  & \begin{CJK*}{UTF8}{gbsn}水果很新鲜, 口感很好吃着非常甜\end{CJK*} (The fruit is fresh, and it tastes delicious and sweet.) \\
\textsc{C-Trm}  & \begin{CJK*}{UTF8}{gbsn}他家的水果挺新鲜, 口感挺值的\end{CJK*} (The fruit in this shop is really fresh, and the taste is worth the price.)\\
\cmidrule(lr){1-2}
\textsc{QualityModel} & \begin{CJK*}{UTF8}{gbsn}超喜欢他家水果, 品质好, 口感很好\end{CJK*} (Really like the fruit in this shop, which is of good quality and tastes well.)\\
\textsc{Creater} & \begin{CJK*}{UTF8}{gbsn}份量很足, 水果新鲜, 口感{\bluehl{\uwave{甘甜很解渴}}}\end{CJK*} (The fruit is a big portion and fresh. It tastes {\bluehl{\uwave{sweet, quenching your thirst}}}.)\\
\bottomrule
\end{tabular}
\caption{Case analysis. Texts in parentheses are the corresponding contents translated to English.}
\label{results:case}
\end{table}

\begin{table}[t]
\centering
\scriptsize
\begin{tabular}{ccc}
\toprule
\textbf{Approach}  & CTR ($\uparrow$) & CPC ($\downarrow$) \\  
\midrule
\textsc{Base}           & - & -\\
\textsc{QualityModel}       & +4.5\% &  -4.1\%\\
\textsc{Creater}        & \textbf{+6.9\%} & \textbf{-6.1\%}\\
\bottomrule
\end{tabular}
\caption{Online results (relative improvement).}
\label{results:online}
\end{table}

\subsection{Online Experiments}\label{experiment:online}
We have deployed \textsc{Creater} to a leading advertising platform. 
Our online experiment is conducted for one-week, and all ads are displayed in mobile news feed. For the ad that containing more than one generated texts (because there may be multiple control codes), we randomly choose one of them to display. 
The experiment traffic covers over 12,000 advertisers, and results are computed based on over ten million impressions to ensure the confidence of online metrics.

We compare performance among the ad texts generated by \textsc{Creater}, \textsc{QualityModel}, and those provided by advertisers (as \textsc{Base}). Core metrics are CTR and cost per click (CPC): CTR $=\frac{\text{\#click}}{\text{\#impression}}$  reveals attractiveness; CPC $=\frac{\text{total cost of advertisers}}{\text{\#click}}$ reflects ad delivery efficiency.   
Table~\ref{results:online} shows that \textsc{Creater} achieves significantly improvements on both CTR and CPC, verifying its effectiveness of improving delivery efficiency.

\section{Related Work}

Most studies focus on generating ad texts given landing page contents~\cite{thomaidou2013automated}. 
~\citet{hughes2019generating} employ a CTR model as reward estimator with self-critical RL, and~\citet{kamigaito2021empirical} consider fluency, relevance and quality rewards to capture the characteristics of effective ad texts.
~\citet{kanungo2021ad} incorporate masked language modeling with self-critical learning to improve the generation for multiple products.~\citet{wang2021reinforcing} design model-based RL system that mimics real user feedback.

To model user click behavior,
~\citet{wang2020evolutionary} take click as a measure of text fitness and design click-based reward.~\citet{wang2019quality} build a CTR space to obtain sample quality that weights cross-entropy loss. 
Unlike these work, we directly model the distinctness of positive and negative targets, and propose a customized pre-training objective.

\section{Conclusion}
We propose \textsc{Creater} for generating ad texts, which employs contrastive learning to encourage the model to generate texts achieving higher CTR. We design a novel self-supervised objective customized to our scenario, reducing the gap to further fine-tuning. Experiments verify that \textsc{Creater}  brings significant uplift on core metrics. 

In future work we will take a next step to improve faithfulness, and extend the model to handle multiple aspects~\cite{chan2021controllable} and multiple reviews (which may be conflicting) with graph neural networks~\cite{wei2021graph}.

\section*{Acknowledgments}
We thank all the anonymous reviewers to their valuable comments for improving this work.

\section*{Ethical Considerations}
When we apply large-scale corpora from the Web, alleviating bias issues is necessary. 
We make efforts from two perspectives: 
(1) For input reviews, we have filtering steps to remove harmful contents, and ensure that they do not have user privacy information like age and gender (``Data Collection and Filtering'' of §~\ref{appendix:datacollection}); 
(2) For output ad texts, we are cautious before online deployment with a risk control procedure (``Post-Processing before Deployment'' of §~\ref{appendix:datacollection}). 
(3) Our model does not use user privacy information like age and gender.

\bibliography{anthology,custom_short}
\bibliographystyle{acl_natbib}

\appendix

\section{Appendix}
\label{sec:appendix}

\subsection{More Details of Dataset Construction}\label{appendix:dataset}
\paragraph{Data Collection and Filtering}\label{appendix:datacollection}
As mentioned in §~\ref{experiment:dataset}, we construct the dataset $\mathcal D$ where each sample is a tuple of (user review, aspect term, positive ad text, negative ad text). 
The construction procedure of $\mathcal D$ mainly contains the following steps: 
\begin{enumerate}[1)]
    \item Collecting a set of high-quality user reviews $\mathcal D_x$. Firstly, a large size of reviews of e-commerce and retail items are collected. We then filter out low-quality ones via a set of rules (e.g., length constraint, repeat term constraint and harmful/abusive word vocabulary) and a spam detection model (trained based on both text contents and fraud behavior features). After this step, we obtain a review corpus $\mathcal D_x$ containing 1,471,106 reviews, which is also utilized to pre-training. 
    \item Building an aspect term set $\mathcal D_c$, utilized to guide generation and ensure the relevance between review contents and ad texts. According to business demands, we first construct a seed set provided by advertisers. We then expand this small set via an unsupervised extraction model \textsc{Abae}~\cite{he2017unsupervised}, trained on the review corpus  $\mathcal D_x$. Each aspect term is typically represented as a word. After a simple filtering rule based on IDF to remove noise, we obtain an aspect term set $\mathcal D_c$ containing 991 terms. 
    \item Professional editors write two distinct ad texts for each given (user review, aspect term) pair. Because writing high-quality ad texts is time-consuming and labor-intensive, this procedure collects around 50,000 samples. We check the correlation between input and output via randomly sampling a fraction of all tuples written by the same editor, and remove low-quality ones. Besides, we ensure that in a paired sample the ad text does not match word-for-word to the original review. 
    \item Conducting online A/B test to collect user preference (i.e., CTR) on these ad texts. Traditionally, advertisers resort to this step to polish their ad texts for catching user interests. In this work we make use of these data to train contrastive learning based generation model. 
    \item Filtering out invalid tuples to obtain the final dataset $\mathcal D$. We remove outlier samples during online A/B test, e.g., the ads that do not have sufficient impressions or obtain anomalously high CTR. We also use Z-test to ensure that the CTR difference between two ad texts of same ad is significant. As a result, this dataset contains 43,985 samples and covers 4,047 advertisers.%\footnote{Note that we find that the positive/negative ad texts can be distinguished by a fine-tuned \textsc{Bert} classification model~\cite{devlin2019bert}, thus there exist.} 
\end{enumerate}

No  personal identifiable information is included in our dataset: (1) During collection, only review texts are saved, and other meta-information (such as original authors) is not collected. (2) To exclude identifying information which may be contained in texts, we employ regular expression for replacement by placeholders. 

\paragraph{Post-Processing before Deployment}\label{appendix:riskcontrol}
Before online deployment, we have a risk control procedure to cautiously perform post-processing on the ad texts generated by models, aiming to ensure the suitability of ad texts before displaying. For instance, text contents that contain false, useless or harmful information cannot be displayed to users. 
Specifically, this procedure removes the texts containing non-compliant words (e.g., harmful words), and performs manual-checking on generated texts. Overall, the passing rate of generated texts is around 90\% to 95\%, which means that the generation models can be deployed online in industry. 
\end{document}